\documentclass[journal]{IEEEtranTIE}
\usepackage{graphicx}
\graphicspath{{./fig/}}
\usepackage{epstopdf}
\usepackage{cite}
\usepackage{picinpar}
\usepackage{amsmath,amsfonts,amssymb}
\usepackage{url} 
\usepackage[latin1]{inputenc}
\usepackage{colortbl}
\usepackage{soul}
\usepackage{multicol, multirow}
\usepackage{pifont}
\usepackage{xcolor}
\usepackage{alltt}
\usepackage[hidelinks]{hyperref}
\usepackage{enumerate}
\usepackage{siunitx}
\usepackage{breakurl}
\usepackage{pbox}

\usepackage[caption=false,font=footnotesize]{subfig}

\usepackage{lipsum}
\usepackage{xcolor}
\usepackage{tikz}

\newcommand{\JLEEcolor}{black}				
\newcommand{\jrev}{\color{\JLEEcolor}{}} 	
\newcommand{\CLEEcolor}{black}				
\newcommand{\crev}{\color{\CLEEcolor}{}} 	
\newcommand{\Sehooncolor}{black}				
\newcommand{\orev}{\color{\Sehooncolor}{}} 	

\begin{document}
\title{Towards Accurate Force Control of\\Series Elastic Actuators Exploiting a Robust Transmission Force Observer}
\author{
	Chan Lee$^{1}$,
	Jinoh Lee$^{2}$,	and 
	Sehoon Oh$^{1}$ 
	\thanks{
		
		$^{1}$The authors are with Department of Robotics Engineering, Daegu Gyeongbuk Institute of Science and Technology(DGIST), Daegu, Korea, 711-785.
    	$^{2}$Jinoh Lee is with Department of Advanced Robotics, Istituto Italiano di Tecnologia (IIT), Via Morego 30, 16163, Genova, Italy.
    	 (\textit{Chan Lee and Jinoh Lee are co-first authors.})				
	}
}
\maketitle

\begin{abstract}
This paper develops an accurate force control algorithm for {\jrev series elastic actuators} (SEAs) based on a novel force estimation scheme, called transmission force observer (TFOB). {\jrev The proposed method} is designed to improve an inferior force measurement of the SEA caused by nonlinearities of the elastic transmission and measurement {\jrev noise and error of its} deformation sensor. This paper first analyzes the limitation of the {\jrev conventional methods for the SEA transmission force sensing and then} investigates {\jrev its stochastic characteristics, which indeed provide the base to render the accurate force control performance incorporated with the TFOB.  
In particular, a tuning parameter is introduced from holistic closed-loop system analyses in the frequency domain. This gives a guideline to attain optimum performance of the force-controlled SEA system. The proposed algorithm is experimentally verified in an actual SEA hardware setup.}
\end{abstract}
\vspace{-1em}
\section{Introduction} \label{sec:intro}
A {\jrev highly accurate} force control of an actuator is {\jrev a} core technology for {\jrev the modern} mechatronic systems, robotics, and industrial applications which enables {\jrev dynamic control with high fidelity allowing} not only rapid tracking performance but also compliant behavior. In particular, a collaborative {\jrev robot, which is one of} key drivers in recent technologies such as Industry 4.0, essentially {\jrev requires the precise force control capability in its actuation}.
{\jrev This has also promoted the development of force sensing methods such as sensor-based measurement or sensorless estimation in the force-controlled actuator system.}

{\jrev Series elastic actuators} (SEAs) have 
{\jrev facilitated the} aforementioned applications owing to {\jrev its inherent capability} to control output force without {\jrev an extra} force sensor or estimation {\jrev algorithms~\cite{lee2017generalization}. 
	The} SEA employs an elastic element instead of a stiff mechanical structure of the force sensor and the {\jrev intrinsic} compliance leads to both superior force control and safe interaction, {\jrev where} the force measurement can be {\jrev acquired through} 
strain-stress phenomenon. 
{\jrev Accordingly, the precision and the robustness of force control in the SEA system become highlighted as an essential issue~\cite{7579567} aiming at not only high force fidelity, but also dynamic motion with compliant behaviors in high-level robot control applications~\cite{tsagarakis2017walk, 491410}.}

\begin{figure}[t!]
	\centering
	{\includegraphics[trim= 0.0px 0.1in 0in 0.1in, clip,width=0.9\columnwidth]{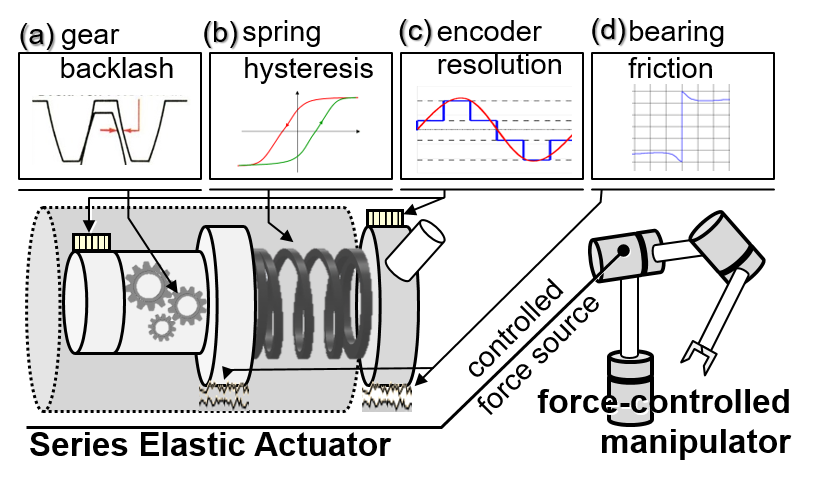}}
	\caption{The sources of inaccuracy in the force measurement and control of SEAs: (a) the gear transmission such as a backlash, (b) mechanical behavior of the spring, (c) errors in encoder measurement signals, and (d) the frictions.}
	\label{fig:intro_SEA}
	\vspace{-1em}
\end{figure}

{\jrev For the force-controlled SEA, the force in the elastic transmission is generally given by the assumption that the transmission force can be simply calculated from a linear spring model and its deflection measured by encoders~\cite{negrello2017design}.}
{\jrev Ideally, the SEA with this deflection-based transmission force sensing can offer} 
better force control performance than 
{\jrev that with the load cell type measurement. However, it} 
suffers from {\jrev considerable} inaccuracy caused by 
{\jrev the inherent nonlinearities in the components in the SEA such as backlash and friction effect in a gear train and the hysteresis in strain-stress characteristics of the spring element, as illustrated in Fig.~\ref{fig:intro_SEA}.}  
{\jrev It is also known that the low resolution of the encoders and noises in the deflection measurement} 
result in performance degradation. 

{\jrev To mitigate these issues, significant efforts have been made in two folds: 1) control design approaches and 2) observer design approaches.} 
{\jrev First, in the control design-oriented approaches, a common remedy is to compensate the measurement inaccuracy by using a}  
lookup 
{\jrev table}
between the encoder measurement of SEA and 
{\jrev the force measured by a} load cell {\jrev as its ground truth}~\cite{wang2015novel}. 
This approach can be 
{\jrev applied}
in a practical way when the measurement errors are deterministic in {\jrev the operating positions of the SEA}.
Wang et al.~\cite{wang2014output} and Choi et al.~\cite{choi2017low} have analyzed the hysteresis effect of {\jrev the} spring in the SEA, and have proposed {\jrev compensation methods based on hysteresis models.} 
On the other hand, 
{\jrev this problem was also tackled by applying robust controllers.} Makarov et al. {\jrev proposed an} 
H$_\infty$ control method 
{\jrev for the elastic joint robot} with uncertainties~\cite{makarov2016modeling}. 
{\jrev In~\cite{7579567},} 
Oh and Kong developed {\jrev a disturbance observer (DOB)-based} 
controller for SEAs which can reject the external (load side) and the internal (spring and motor sides) {\jrev disturbances.} 

{\jrev Second, there have been model-based observer approaches to fundamentally enhance the SEA sensing capability for better force control.} 
Austin et al. adopted {\jrev a Luenberger observer for state estimation of 
	the SEA equipped with the nonlinear} rubber spring~\cite{austin2015control}. Extending 
{\jrev this idea benefiting from the use of} SEA dynamics to overcome the measurement issues, some research groups have {\jrev exploited} 
the motor-side dynamic model and measurement as {\jrev an alternative to} 
the spring deformation-based force measurement. This method has been implemented in {\jrev forms} 
of the residual-based observer~\cite{ruderman2016compensation, lee2017two} and the disturbance observer~\cite{mitsantisuk2013design}, {\jrev and further improved by using} 
both motor side dynamics and spring information~\cite{lee2018residual, yamada2018proposal}.

{\jrev Interestingly, it is noted that the research has been rarely conducted to incorporate the force controller and the model-based observer for the SEA, in spite of potential advantages expected from both approaches.} 
{\jrev One} example is the sliding mode control method for the position control of an elastic joint, proposed in~\cite{ruderman2016sensorless}, which utilized 
{\jrev the} residual-based force observer using motor-side dynamics to overcome hysteresis and friction effects. 
{\jrev Although} the model-based observer is successfully {combined with the position controller of the elastic joint,} 
it is difficult to extend the control methodology directly to the SEA force control. 
{\jrev Hence, designing force control with transmission force estimation techniques} 
{\jrev is still challenging, yet worthwhile to attain accurate and robust performance in SEAs.} 


{\jrev This paper thus} aims to develop {\jrev an SEA force controller with high precision} 
taking full advantage of a robust force estimation method to overcome defective force measurement. The proposed 
{\jrev algorithm} is designed {\jrev through analyses of the measurement error characteristics of the SEA, and} 
quantitatively optimized 
{\jrev with the consideration of the dynamics, the controller and the observer of the entire closed-loop system.}

{\jrev The rest of the paper is organized as follows: first, 
	the errors in the SEA force measurement and their characteristics are mathematically modeled and analyzed} 
in {\crev Section}~\ref{sec:force_estimation}. Section~\ref{sec:force_observer} {\jrev presents the design and verification of a} 
novel force observer {\jrev for SEAs, named transmission force observer (TFOB).} 
{\jrev With exploitation of the TFOB, a force control algorithm} 
is proposed in {\crev Section}~\ref{sec:force_control}. 
{\jrev Particularly, a systematic tuning method to obtain the accurate and robust force control performance is given based on the insight from error analyses.}
Section~\ref{sec:experiment} {\jrev experimentally verifies the holistic force control algorithm with TFOB in the real SEA test bench.}

\section{Problems in Force Measurement of SEAs}\label{sec:force_estimation}
To deal with the accurate force estimation/control problem, inaccuracy issues of {\orev the conventional deformation-based} force measurement in SEA are analyzed in this section.
Firstly, the mathematical model for deformation-based force measurement is defined to confirm  error factors which are discussed in this paper. In order to explore the behavior of the errors, two types of experimental analyses are conducted. One is deterministic analysis, and the second is stochastic analysis.
The deterministic analysis verifies the non-linear characteristic of deformation-based force measurement, and stochastic analysis indicates that {\jrev the nonlinear characteristic of the SEA force measurement error is regarded as Gaussian noise. The} 
Gaussian characteristic of the error {\jrev will provide a connection to the} 
performance optimization of {\jrev the} force controller design in {\crev Section}~\ref{sec:force_control}.
\vspace{-1em}
\subsection{Errors in Deformation-based Force Measurement}\label{sec:force_estimation_issue}
Conventional force/torque measurement of SEA, 
{\jrev called} deformation-based force measurement (DFM), is usually given as follows:
\begin{equation}
\hat{\tau}^s_s = K^n_s \theta^m_s, \label{eq:deformation_force}
\end{equation}
where $\hat{\tau}^s_s$ is the estimated spring force, $K_s^n$ is the nominal spring stiffness, and $\theta^m_s$ is the measured spring deformation by encoders. Note that the term {\jrev `force' is used for general explanations hereinafter, while the term `torque' is used for one regarding the experimental results since a rotary-type SEA is set as the experiment hardware.} 

{\jrev However, in this DFM, the accuracy is often deteriorated because} 
$\hat{\tau}^s_s$ is subject to {\jrev error factors such as} 
encoder resolution, noise,  backlash in gears and uncertainty in spring behavior 
{\jrev as depicted} in Fig.~\ref{fig:intro_SEA}. These defective factors can be categorized as two folds: $\theta_s^e$, the error of deformation measurement itself (resolution and noise problem), and $K_s^e$, the errors in estimation model (spring hysteresis, friction and backlash). 

{\jrev The DFM}~\eqref{eq:deformation_force} can be {\jrev rewritten with consideration of the error factors $K_s^e$ and $\theta_s^e$ as} 
\begin{eqnarray}
\hat{\tau}^s_s &=& {\jrev (K_s + K_s^e )} \left(\theta_s+\theta_s^e\right), \nonumber \\
&=& K_s\theta_s + K_s^e\theta_s+K_s\theta_s^e+ K_s^e\theta_s^e , \label{eq:est_error}
\end{eqnarray}
where $K_s\theta_s$ is actual force exerting from the spring, and the terms $K_s^e\theta_s$, $K_s\theta_s^e$ and $K_s^e\theta_s^e$ represent errors. The errors in \eqref{eq:est_error} can be {\jrev rearranged as follows: 
	\begin{eqnarray}
	\hat{\tau}^s_s &=& \tau_s + K_s\left(\frac{K_s^e}{K_s}\theta_s+\frac{K_s^n}{K_s}\theta_s^e\right), \nonumber \\
	&=& \tau_s + K_s\left(\xi_s^m+\xi_s^e\right), \nonumber\\
	&=& \tau_s + K_s\xi_s^* , \label{eq:DFM_tau}
	\end{eqnarray}
	
	\noindent where} $\xi_s^*$ is a total measurement error including mechanical error $\xi_s^m$ and encoder measurement error $\xi_s^e$. The influence of $\xi_s^*$ on the force measurement $\hat{\tau}^s_s$ changes depending on the
level of $K^n_s$; the larger the spring stiffness is, the more affected by the measurement noise the force measurement is.

\begin{figure}
	\centering
	\subfloat[]{\includegraphics[width=0.54\columnwidth]{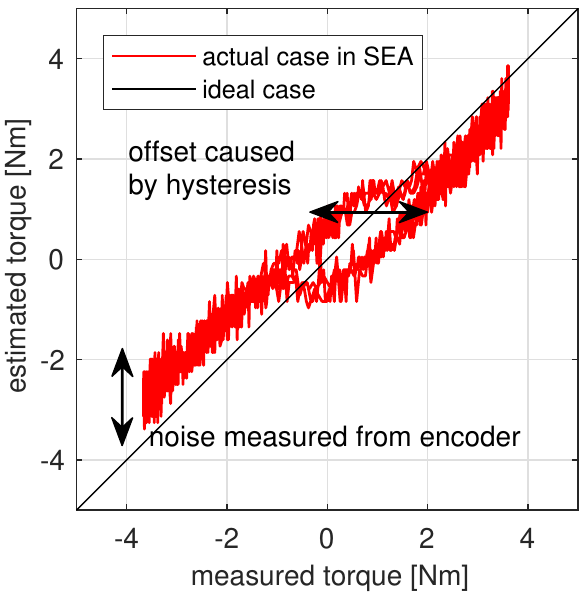}}~
	\subfloat[]{\includegraphics[width=0.45\columnwidth]{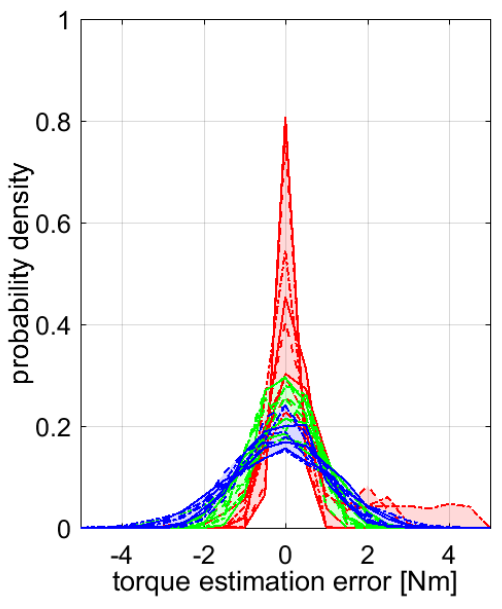}}
	\caption{{\jrev Plots of torque estimation errors caused by nonlinear characteristics:} 
		(a) experimental analyses of deterministic behavior and (b)~stochastic behavior, where  blue, green and red colors indicate high (9100~Nm/rad), mid (2500~Nm/rad) and low (60~Nm/rad) stiffness values.}
	\vspace{-1em}
	\label{fig:offset_error}
\end{figure}

{\crev To further investigate this problem, we perform force measurement experiments, where} 
the torque output of {\jrev the rotary-type} 
SEA was measured based on DFM in
\eqref{eq:deformation_force} and compared with {\jrev the torque reading from an additional torque sensor equipped as the ground truth. (Refer to the details of the experimental setup in {\crev Section}~\ref{sec:setup}.)} 
%
{\crev Fig.} \ref{fig:offset_error}(a) shows the result of comparison between the DFM of SEA and actual torque. {\jrev One can clearly observe the}  
discrepancy between the DFM and {\jrev ground-truth torque,} 
while {\jrev their trends are well-matched.} 
{\jrev This discloses two issues: 1) the offset of 2~Nm (in particular, when the direction of the force changes), mainly due to the hysteresis of the spring or the backlash among gears; and 2) the noises. The former error corresponds to $\xi_s^m$, which is caused by the mechanism, while the latter error corresponds to $\xi_s^e$.} 
\vspace{-1em}
\subsection{Stochastic Characteristics of {\jrev the} Measurement Errors}
{\orev The measurement error $\xi_s^*$ of DFM exhibits stochastic characteristics, which can be apparently verified with measurement experiments under regulated pattern torque generation. For analysis of this characteristic, experiments have been conducted, where the SEA is controlled to generate several sinusoidal pattern torque outputs.}

{\orev In the experiment, a torque sensor is connected between an SEA and a fixed environment to measure the accurate torque output of SEA. For in-depth analysis of the measurement error under various operating condition, 9 types of sinusoidal torque references (3 different frequencies $\times$ 3 different magnitudes) and three types of spring settings (low, mid and high stiffness) were tested in the experiments.} 
\noindent In each {\jrev experiment,} {\orev the torque outputs of SEA were estimated by DFM \eqref{eq:deformation_force} and compared with the torque sensor measurements. The difference between two was calculated, and Fig.~\ref{fig:offset_error}(b) illustrates the distribution of these differences, which are the estimation error by DFM.}



From the error distributions, one can notice that the torque error measurement of SEA is regarded to have Gaussian distribution. {\orev The comparison among three different spring stiffness verifies that large spring stiffness $K_s$ leads to large torque estimation error.}


It is well known that the encoder quantization error can be modeled as Gaussian noise, and the result verifies that the torque estimation is affected by this Gaussian encoder noise. This implies that the low resolution encoder with high stiffness spring can lead to very inaccurate force estimation.

This observation shows that the DFM error factor $\xi^*_s$ can be considered Gaussian, the magnitude of which can be evaluated using the variance. 
This point can be utilized in the feedback controller design to quantitatively evaluate controllers.

\section{Robust Transmission Force Observer {\jrev (TFOB)}}
\label{sec:force_observer}

\subsection{Design of the {\jrev TFOB}}
\label{sec:observer_theory}

{\jrev In this section, an observer to estimate the SEA force output is proposed to address this inaccurate force estimation.} 
The observer exploits the {\jrev fact} 
{\orev that the SEA output force works as the external disturbance (reacted spring force) with respect to the motor dynamics, which allows for utilization of force observer concept~\cite{haddadin2017robot, murakami1993torque, oh2014design} to observe the SEA output force as the external disturbance.}
\begin{figure}
	\centering
	\includegraphics[trim= 0in 0in 0.1in 0in, clip, width=0.85\columnwidth]{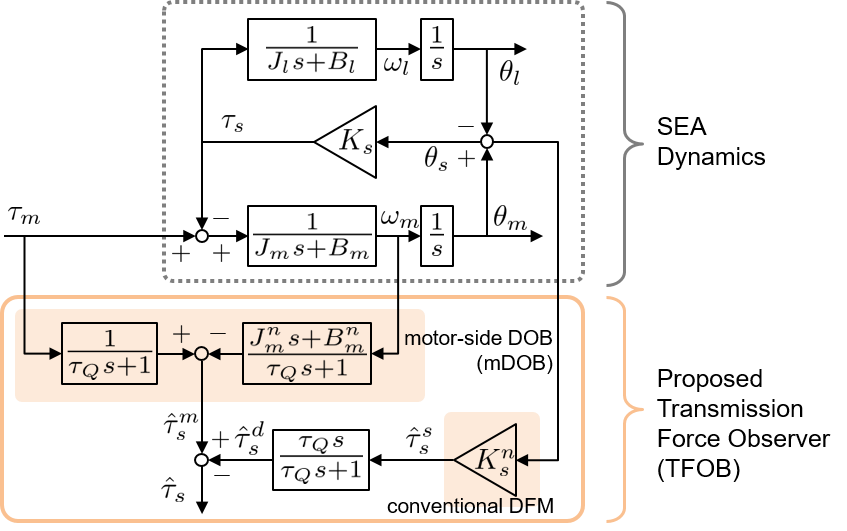}
	\caption{{\jrev Block diagram of the proposed TFOB for SEAs.}}
	\label{fig:observers_SEA}
	\vspace{-1em}
\end{figure}

{\crev Fig.} \ref{fig:observers_SEA} shows the dynamics of SEA with the force observer design; the SEA dynamics in the upper block diagrams consists of the motor dynamics $P_m(s)=\frac{1}{J_ms + B_m} $, the spring $K_s$ and the
load dynamics $P_l (s)=\frac{1}{J_ls + B_l}$, where $J_{\bullet}$ and $B_{\bullet}$ represent inertia and damping values of motor-and load-sides. The output force of SEA is presented as $K_s \theta_s$.

A motor-side DOB (mDOB) of SEA is designed by utilizing the nominal motor dynamics $P^n_m(s)=\frac{1}{J_m^ns + B_m^n}$, which can estimate the external force at the motor-side. In this case, the spring force $K_s \theta_s$ corresponds to the external force.

The estimate of the spring force by mDOB can be formulated as {\jrev follows:} 
\begin{equation}
\hat{\tau}^m_s = - Q(s) \left( {P_m^n}^{-1}(s) \omega_m^m - \tau_m \right), \label{eq:dob_estimate}
\end{equation}
where {\jrev  $Q(s)$ denotes the Q filter expressed as a form of the low-pass filter--- $Q(s)=\frac{1}{\tau_Qs+1}$ in this paper, and $\omega^m_m$ denotes the motor angular velocity measured as follows:} 
\begin{equation}
\omega^m_m= s\left(\theta_m+\xi_m^*\right), \label{eq:encoder_noise_m}
\end{equation}
where  $\xi^*_m$ represents the measurement error of the motor encoder caused by mechanical {\orev quantization or noise}, and thus $\omega_m^m$ is influenced by numerical differentiation of measurement errors. 
{\jrev Note that the Q filter $Q(s)$} 
should be added to reduce this error, {\jrev whereas it also limits} 
the estimation performance only within the bandwidth of $Q(s)$, 
{\jrev where the cut-off frequency 
	is denoted as $\omega_{Q}=1/\tau_Q$.} 

Accordingly, the spring force can be estimated in two ways: $\hat{\tau}^s_s$ by DFM as in \eqref{eq:deformation_force} and $\hat{\tau}^m_s$ by the above mDOB. Each estimate has its own drawback: $\tau_s^s$ has offset and noise problem and $\tau_s^m$ has bandwidth limitation. 

To overcome these drawbacks, this paper proposes a novel algorithm to integrate two estimates in a complimentary way. The integrated transmission force observer (TFOB) to achieve accurate force estimation is designed as follows.
\begin{eqnarray}
\hat{\tau}_s &=& \hat{\tau}^m_s + (1-Q(s))\hat{\tau}^s_s = \hat{\tau}^m_s + \hat{\tau}^d_s\label{eq:tfob} \nonumber \\ 
&=& Q(s) \left( \tau_m -{P_m^n}^{-1}(s) \omega_m^m \right) + (1-Q(s)) K^n_s \theta^m_s \nonumber \\ 
&=& \tau_s+\left(Q(s){P_m^n}^{-1}(s)s\xi_{m}^*- (1-Q(s))K_s^n\xi_{s}^*\right) \label{eq:TFOB}
\end{eqnarray}

The conventional force estimation $\tau_s^s$ is high pass-filtered in the proposed TFOB as shown in \eqref{eq:TFOB}. The offset issue in $\tau_s^s$ can be addressed by this high pass filtering. The cut-off frequency of $Q(s)$ is the tuning factor of TFOB, which determines the bandwidths of mDOB-based estimation and high pass-filtered DFM. {\jrev Note that the baseline of TFOB is inspired from~\cite{8371176}, however, it is redesigned in a different form with consideration of} 
the motor-side encoder noise to extend the analysis to a controller optimization problem in {\crev Section}~\ref{sec:force_control}.

%
%
%
%
%
\vspace{-0.8em}
\subsection{Verification of Estimation Accuracy of Proposed TFOB}\label{sec:estimation_verification}
\begin{table}[t]
	\centering
	\caption{RMS errors of spring force estimation methods}\label{tab:std_est}
	\begin{tabular}{r c c c c }
		\hline \hline
		$\omega_Q$&(a) 0.1~Hz&(b) 1~Hz&\textbf{(c) 5~Hz}&(d) 10~Hz \\
		\hline
		TFOB-based  &0.678~Nm& 0.311~Nm& \textbf{0.298~Nm}& 0.403~Nm \\
		DFM-based &	0.770~Nm& 0.791~Nm& 0.783~Nm& 0.817~Nm	\\
		\hline \hline
	\end{tabular}
	\vspace{-1em}
\end{table}
\begin{figure}[t]
	\centering
	\subfloat[]{\includegraphics[ width=0.24\columnwidth]{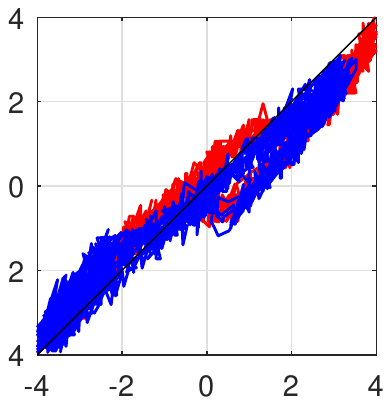}}~
	\subfloat[]{\includegraphics[ width=0.24\columnwidth]{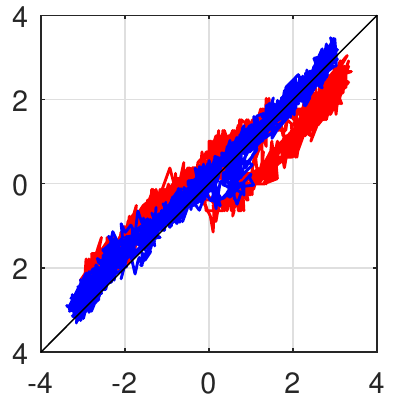}}~
	\subfloat[]{\includegraphics[ width=0.24\columnwidth]{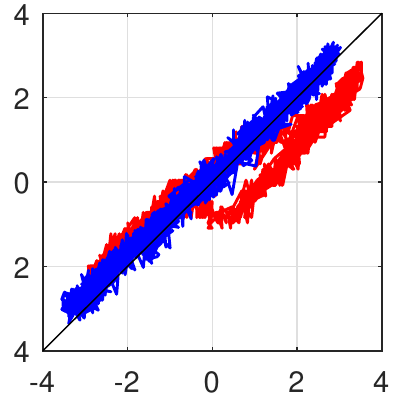}}~
	\subfloat[]{\includegraphics[ width=0.24\columnwidth]{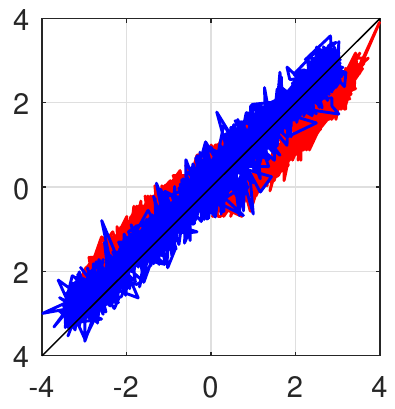}}
	\caption{TFOB performances with regards to varying Q filter bandwidth, $\omega_Q$: (a) 0.1~Hz, (b) 1~Hz, (c) 5~Hz, and (d) 10~Hz. Red and blue lines indicate the results of DFM and TFOB respectively. $x$-axis is {\jrev the} estimated torque, and $y$-axis is {\jrev the} measured torque [Nm].}
	\label{fig:est}
	\vspace{-1em}
\end{figure}

In order to verify the performance of TFOB, the estimation experiments were performed, and the results were compared with the torque sensor measurement. In particular, the experiments were conducted using various Q filter bandwidth settings of TFOB, {\orev in search of} the optimal bandwidth value. This will give a notion that the Q filter bandwidth can be further exploited to achieve optimal force control performance (discussed in {\crev Section}~\ref{sec:force_control}).

{\crev Fig.}~\ref{fig:est} shows results of the estimation by TFOB compared with conventional DFM estimate. In the experiments, the bandwidths of Q filters $\omega_Q$ varies from (a) 0.1~Hz  to (d) 10~Hz. In all subplots in Fig.~\ref{fig:est}, the red lines indicate results of DFM, and the blue lines indicate results of TFOB. 

DFM results show large off-sets up to 1~Nm, even though it exhibits good linearity. In contrast, TFOB can successfully reduce the error in all the results. Comparison of (a) to (c) reveals that high $\omega_Q$ can reduce the error more effectively. However, when the bandwidth of the TFOB increases to~10 Hz, the noise increases in the estimate and shows worse estimation result than 5~Hz. This is due to error in the measurement of the motor-side encoder, $\xi_m^*$.

In order to quantitatively show these trade-off characteristics, the Root Mean Square Error (RMSE) values of the estimation are compared in Table~\ref{tab:std_est}. The result shows that the TFOB with 5~Hz bandwidth shows the best performance.

\section{Accurate Force Control Based on TFOB}\label{sec:force_control}
{\orev In this section, design method of force controllers utilizing TFOB is proposed, and its characteristic is analyzed in terms of accuracy or robustness against measurement errors. Finally, a methodology to tune Q filter of TFOB is discussed taking into consideration the dynamic characteristics of TFOB-based force control.
}
\subsection{Design of TFOB-based Force Controller}

{\orev TFOB does not conform the force controller of SEA, in other words, any type of controller that has been proposed for high performance control of SEA force output can}
benefit from the TFOB by replacing the conventional force estimate $\hat{\tau}^s_s$ with the TFOB output $\hat{\tau}_s$. The control law of the proposed TFOB-based force control is designed {\jrev as
	{\orev \begin{align}
		\tau_m = C&_f(s)\left(\tau_s^r-\hat{\tau}_s\right), \nonumber\\
		= C&_f(s)\{\tau_s^r-Q(s)\left(\tau_m -P_m^{-1}(s) \omega_m \right) \nonumber\\
		-& \left(1-Q(s)\right) K_s \theta_s\} .
		\end{align}}}

\begin{figure}[t]
	\centering
	{\includegraphics[trim= 0in 0.15in 0.1in 0in, clip, width=0.8\columnwidth]{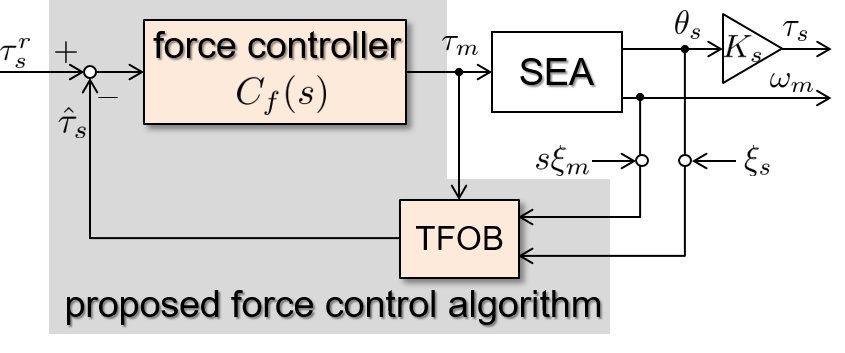}}
	\caption{\jrev The block diagram of the proposed force control with the TFOB.}
	\label{fig:DFC}
	\vspace{-1em}
\end{figure}

{\crev Fig.} \ref{fig:DFC} illustrates the proposed control configuration utilizing feedback from the TFOB output{\orev , where the force controller $C_f(s)$ can be designed as any types of controllers}, e.g., {\jrev proportional-integral-derivative (PID) control, DOB, or sliding-mode control.} 

{\jrev In this paper, the conventional PD controller is employed for  the force controller $C_f(s)$,} 
as it is the most general and widely-utilized control design methodology. In the following subsections, two aspects are to be examined: 1) how the TFOB can improve the force control performance and 2) how to tune the Q filter when a controller is given.

\subsection{Open-loop Analysis of SEA Dynamics}\label{sec:SEA_dyn}
\begin{figure}
	\centering
	\subfloat[]{\includegraphics[width=0.43\columnwidth]{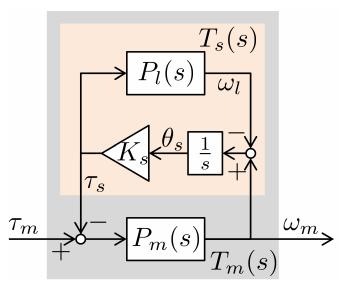}}~
	\subfloat[]{\includegraphics[width=0.51\columnwidth, trim={0in -0.1in 0.1in 0.1in}, clip]{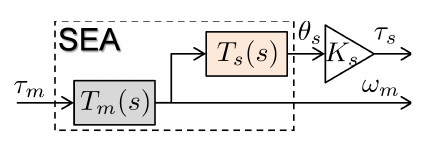}}
	\caption{\jrev Block diagram of the dynamic model of the SEA.}
	\label{fig:block}
	\vspace{-1em}
\end{figure}

To design and analyze TFOB-based force controller, the dynamics of SEA is
investigated at first. {\crev Fig.} \ref{fig:block} re-illustrates the block
diagram of an SEA in Fig. \ref{fig:observers_SEA} with the motor $P_m(s)$,
the spring $K_s$ and the load $P_l(s)$ dynamics.

As two outputs, $\theta_s (\propto \tau_s)$ and $\omega_m$ are utilized for TFOB, the transfer function from the motor torque $\tau_m$ to these two outputs need to be derived as {\jrev follows:}

\begin{eqnarray}
\frac{\theta_s}{\tau_m} \!\!
&\!\!=\!\!&\!\! \frac{P_m(s)}{s+K_s \left(P_m(s)+P_l(s) \right)} \label{eq:TF_sm} \\
\frac{\omega_m}{\tau_m} \!\!&\!\!=\!\!&\!\! \frac{P_m(s)(s+K_sP_l(s))}{s+K_s \left(P_m(s)+P_l(s)\right)} = T_m(s). \label{eq:TF_mm}
\end{eqnarray}

From these transfer functions, the relationship between two outputs $\theta_s$ and $\omega_m$ is derived as

\begin{equation}\label{eq:Ts_relation}
\theta_s = \frac{1}{s+K_sP_l(s)} \omega_m 
= T_s(s) \omega_m.
\end{equation}
From the viewpoint of motor dynamics $P_m(s)$, $\tau_s = K_s T_s(s) \omega_m$
is considered external force, which forms a feedback loop through
$P_m(s)T_s(s)K_s$. This relationship finalizes the transfer function from the
motor torque to the motor angular velocity re-organized using $T_s(s)$ as
\begin{equation}
\frac{\omega_m}{\tau_m} = T_m(s) = \frac{P_m(s)}{1+K_sP_m(s)T_s(s)}.
\end{equation}

In the same way, the transfer function to the force output of SEA $\tau_s$ is
derived as
\begin{equation}
\frac{\tau_s}{\tau_m} = K_sT_s(s)T_m(s) = \frac{K_sP_m(s)T_s(s)}{1+K_sP_m(s)T_s(s)}.
\end{equation}

\noindent These transfer functions are utilized for synthesis and analysis of controller in the following {\jrev subsection}. 
\vspace{-0.8em}
\subsection{Closed-loop Analysis of TFOB-based Force Control}\label{sec:single_feedback}

The reference tracking performance of the proposed control in Fig.
\ref{fig:DFC} is analyzed using the transfer function from the reference
$\tau_s^r$ to the output $\tau_s$, which is given as {\jrev follows:}
\begin{equation}\label{eq:refT_single}
\frac{\tau_s}{\tau_s^{r}} = P_{cl}=  \frac{C_fK_sP_mT_s}{1+(1+C_f)K_sP_mT_s}
\end{equation}

\noindent {\jrev Note that the transfer function (\ref{eq:refT_single}) is the same as that of} 
the conventional DFM-based force control. In other
words, the TFOB does not affect the reference tracking characteristic, {\jrev i.e.,} 
the force controller $C_f(s)$ can be designed independently from TFOB. For simplicity of description, {\jrev hereinafter}, the Laplace domain operator (s) of the system is omitted.

{\jrev The impacts on the control performance from
	the encoder measurement error $\xi_s^e$ and mechanical modeling error $\xi_s^m$---collectively expressed as $\xi_s$ given in \eqref{eq:DFM_tau}--- can be reduced by TFOB. It can be investigated by} 
the transfer functions from $\xi_m$ and $\xi_s$ to {\orev $\tau_s$} {\jrev as follows:} 
\begin{align}
\frac{\tau_s}{\xi_s} =& (1-Q)K_sP_{cl} \label{eq:xiST_single},\\
\frac{\tau_s}{\xi_m} =& sQP_m^{-1}P_{cl} \label{eq:xiMT_single}.
\end{align}

\begin{figure}[t]
	\centering
	{\includegraphics[trim= 0in 0.12in 0in 0in, clip, width=0.85\columnwidth]{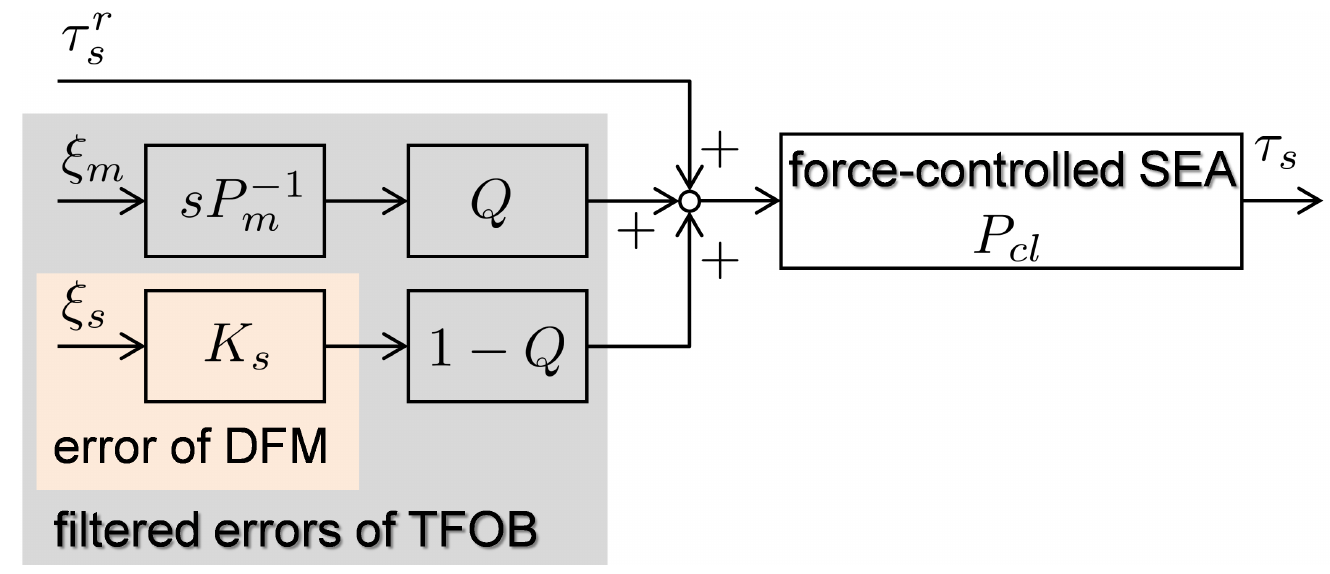}}
	\caption{Influences of errors in {\jrev the closed-loop of DFM- and TFOB-based force control.}} 
	\label{fig:C_TFOB}
	\vspace{-1em}
\end{figure}

\noindent {\jrev Whereas, the impact} 
of $\xi_s$ on $\tau_s$ in the DFM-based force control is given as 
\begin{equation}\label{eq:noise_single}
\frac{\tau_s}{\xi_s} = K_sP_{cl}.
\end{equation}
{\jrev As illustrated in Fig.~\ref{fig:C_TFOB}, it then can be noticed in the TFOB-based control that the effect of ${\xi_s}$ is high-pass filtered by $(1-Q)$ as shown in~\eqref{eq:xiST_single}. Besides,} 
$\xi_m$ {\jrev has no effect on} 
the conventional DFM-based control, {\jrev but affects $\tau_s$ in TFOB-based control which is low-pass filtered by $Q$ shown in~\eqref{eq:xiMT_single}}. 
%
%

{\jrev Accordingly, for} TFOB-based force control to achieve better error reduction, {\jrev magnitudes of the transfer functions~\eqref{eq:xiST_single},~\eqref{eq:xiMT_single}} should be smaller than that of \eqref{eq:noise_single}. 
{\orev {\jrev Henceforth,} the output force caused by the measurement errors under the TFOB-based control is investigated 
	with various conditions and compared with the DFM-based control.} 

To handle the error characteristics of two different encoder measurements ($\xi_s$ and $\xi_m$) in a comprehensive way, two error characteristics are quantified as {\jrev follows:}
\begin{equation}\label{eq:two_noises}
|\xi_m|^2 = H^2 |\xi_s|^2,
\end{equation}
which means the ratio of two error magnitudes {\orev can be related using arbitrary gain $H$. In other words, the difference between measurement error conditions of two angles $\theta_m$ and $\theta_s$ is described by $H$.} 
For example, if the error is considered due to the encoder quantization, $H$ represents the ratio of the encoder resolutions between the spring encoder and the motor encoder. It is noticeable that {\orev the consideration of gear ratio which is necessary when the motor-side encoder is placed before the gear transmission of SEA, can be reflected in $H$, too.}

With the relationship in~\eqref{eq:two_noises}, the effects of the measurement error on the SEA force output under the TFOB-based control can be calculated as
\begin{equation}\label{eq:error_effect}
\tau_s^{e}= \frac{C_f(1-Q)P_m K^2_sT_s\xi_s   + sC_fQ   K_sT_s\xi_m }{1+K_sP_mT_s\{1+C_f\}},
\end{equation}
and its norm (magnitude) is given as
\begin{align}\label{eq:error_TFOB}
\left\lVert\tau_s^{e}\right\rVert &= \left\lVert \frac{C_f(1-Q)P_m K^2_sT_s\xi_s   +  sC_fQ   K_sT_s\xi_m  }{1+K_sP_mT_s\{1+C_f\}} \right\rVert \nonumber \\ 
&\leq \left\lVert \frac{C_f(1-Q)P_m K^2_sT_s }{1+K_sP_mT_s\{1+C_f\}} \right\rVert \left\lVert\xi_s\right\rVert+ \nonumber \\
&\quad   \left\lVert \frac{sC_fQ   K_sT_s }{1+K_sP_mT_s\{1+C_f\}} \right\rVert \left\lVert\xi_m\right\rVert \nonumber \\
&= \left(\left\lVert \frac{C_f(1-Q)P_m K^2_sT_s}{1+K_sP_mT_s\{1+C_d\}} \right\rVert\right.+ \nonumber \\
&\left.\quad \quad \left\lVert \frac{sHC_fQ   K_sT_s }{1+K_sP_mT_s\{1+C_f\}} \right\rVert\right) \left\lVert\xi_s\right\rVert=\left\lVert\tau_s^{TFOB}\right\rVert.
\end{align}
Meanwhile, the same error characteristic with the conventional DFM-based control is given as
\begin{equation}\label{eq:error_DFM}
\left\lVert\tau_s^{DFM}\right\rVert = \left\lVert\frac{C_fP_m K^2_sT_s}{1+K_sP_mT_s\{1+C_f\}} \right\rVert\left\lVert\xi_s\right\rVert.
\end{equation}
By comparing the magnitude of~\eqref{eq:error_TFOB} and~\eqref{eq:error_DFM}, the effects of the measurement errors on the output force with the proposed TFOB feedback control and DFM feedback control can be {\orev derived and compared.} 

\begin{figure}[t]
	\centering
	\subfloat[]{\includegraphics[width=0.49\columnwidth]{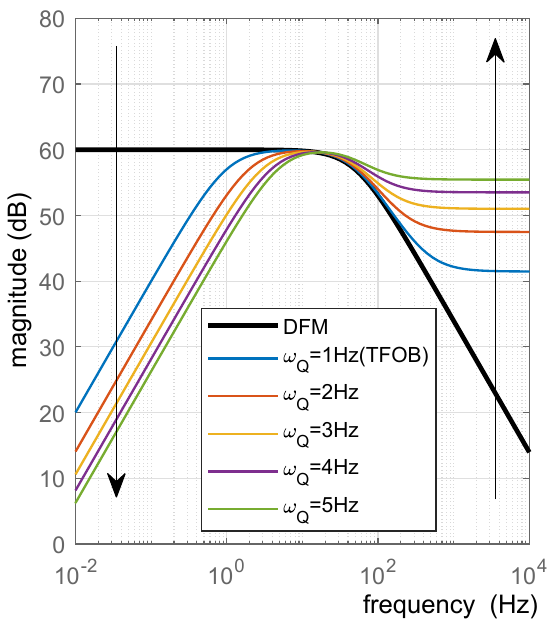}}~
	\subfloat[]{\includegraphics[width=0.49\columnwidth]{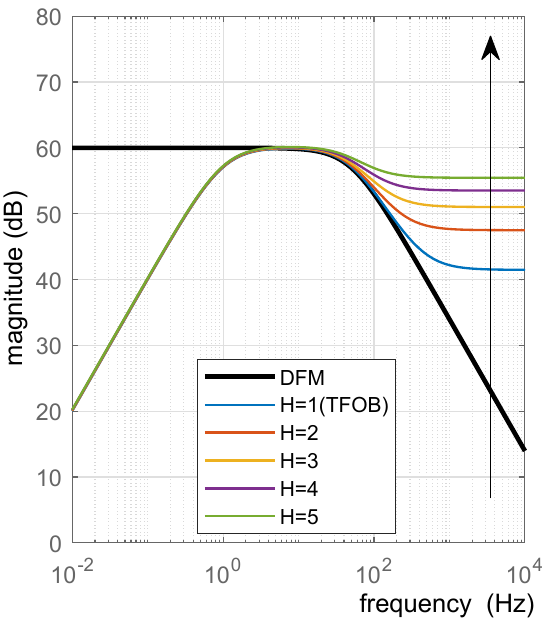}}
	\caption{Influences of measurement errors in SEA force output according to (a) Q filter bandwidth of TFOB, (b) motor encoder performance (higher H means low quality {\jrev encoder.)}}
	\vspace{-1.2em}
	\label{fig:FRF}
\end{figure}

{\jrev In Fig.~\ref{fig:FRF}, the {\orev magnitudes with TFOB feedback in~\eqref{eq:error_TFOB}}
	are calculated with different Q filter bandwidths and $H$ values. As seen in the plots, the magnitude with the TFOB-based control is lower than that with the DFM-based control in the low frequency range, while it becomes higher in the high frequency range. In details, Fig.~}~\ref{fig:FRF}(a) shows that Q filter bandwidth can change the reduction of the magnitude in the low frequency range, while it will sacrifice the high frequency magnitudes. Moreover, it is interesting that $H$ in Fig.~\ref{fig:FRF}(b), the error characteristic of the motor side encoder does not change the low frequency magnitude of the TFOB-based control. Namely, TFOB-based control improves the low frequency error characteristic regardless of motor encoder resolution. But in the high frequency range, the error characteristic is deteriorated as $H$ increases.  


This analysis verifies that the bandwidth frequency of the TFOB Q filter is a tuning factor that adjusts the trade-off between two errors $\xi_m$ and $\xi_s$. The bandwidth of error attenuation can be improved by Q filter bandwidth at the sacrifice of the high frequency error magnitude, which is determined by the characteristic of the motor-side encoder.


\vspace{-0.5em}
\subsection{Tuning method for TFOB-based Control}\label{sec:q_tuning}
\begin{figure}[t!]
	\centering
	\subfloat[]{\includegraphics[trim= 0in 0in 0in 0.7in,width=0.42\columnwidth]{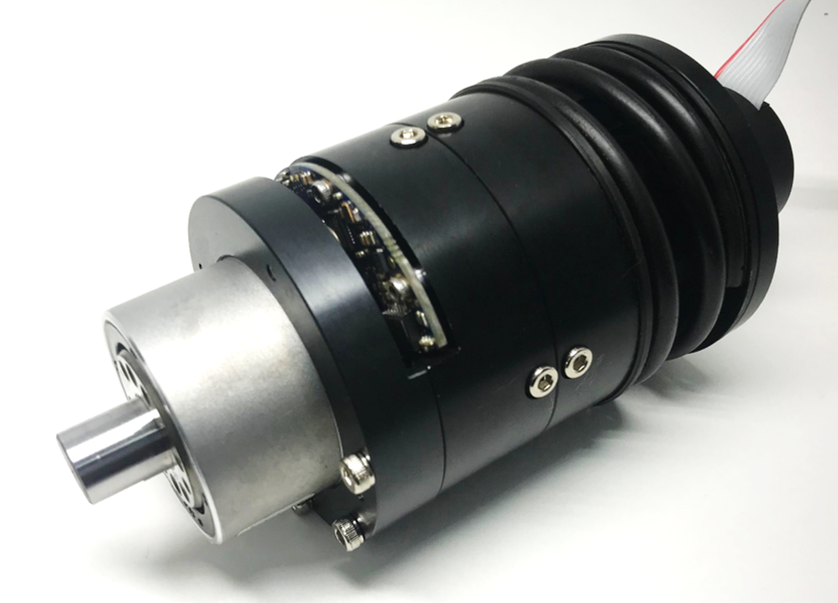}}
	\subfloat[]{\includegraphics[trim= 0in 0in 0in 0.7in,width=0.58\columnwidth]{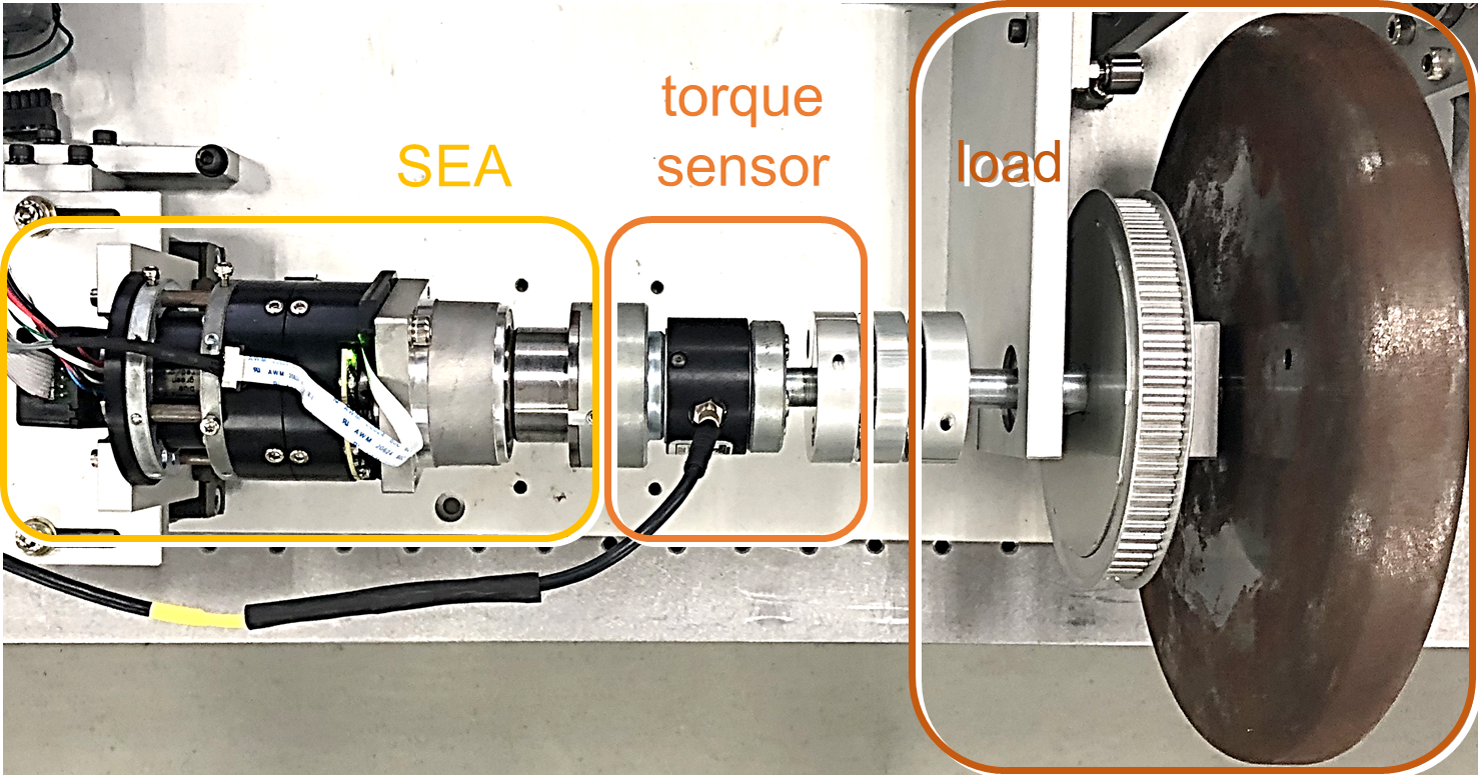}}
	\caption{Experimental setup: (a) the SEA (detailed in ~\cite{7793816}), and (b) test bench for SEA torque control and estimation performance {\jrev verification.}}
	\label{fig:hardware}
	\vspace{-1em}
\end{figure}

Our ultimate goal is to find optimal Q filter bandwidth $\omega_Q$ under the TFOB-based force control, where the condition can be described as follows:
\begin{equation}\label{eq:optim}
\left\lVert \tau_s^{{TFOB}}\right\rVert < \left\lVert \tau_s^{{DFM}}\right\rVert
\end{equation}
For the TFOB-based control to achieve better error reduction performance than the conventional DFM-based force control as described in~\eqref{eq:optim}, the magnitude of~\eqref{eq:error_TFOB} should be less than~\eqref{eq:error_DFM} as {\jrev follows:} 
\begin{eqnarray}\label{eq:error_ieq}
\overbrace{\left\lVert\frac{C_f(1-Q)P_m K^2_sT_s }{1+K_sP_mT_s(1+C_f)}\right\rVert + \left\lVert\frac{ sHC_fQ K_sT_s }{1+K_sP_mT_s(1+C_f)}\right\rVert}^{\text{error {\jrev chracteristics} of TFOB-based control in \eqref{eq:error_TFOB}}} \nonumber\\
<\underbrace{\left\lVert\frac{C_fP_mK_s^2T_s}{1+K_sP_mT_s(1+C_f)} \right\rVert}_{\text{error {\jrev chracteristics} of DFM-based control in \eqref{eq:error_DFM}}}
\end{eqnarray}
\noindent It is difficult to find Q filter condition to satisfying~\eqref{eq:error_ieq} in all frequency ranges, however, the condition to set the infinity norm of two transfer functions as in~\eqref{eq:opt} can be found {\jrev as
	\begin{align}
	\left\lVert \frac{C_f(1-Q)P_m K^2_sT_s }{1+K_sP_mT_s(1+C_f)}\right\rVert_{\infty} + \left\lVert\frac{ sC_fQ K_sT_sH }{1+K_sP_mT_s(1+C_f)}\right\rVert_{\infty} \nonumber\\ 
	< \left\lVert \frac{C_fP_mK_s^2T_s}{1+K_sP_mT_s(1+C_f)} \right\rVert_{\infty}.
	\label{eq:opt}
	\end{align}
	\noindent With} the Q filter design in~\eqref{eq:opt}, the maximum force output error of TFOB-based control is guaranteed to be less than that of DFM-based control as {\orev follows~\cite{sivashankar1992induced}:}
\begin{equation}
\max_{\xi_s(t)}\frac{\left\lVert \tau_s^{{TFOB}}(t)\right\rVert_{2}}{\left\lVert \xi_s(t)\right\rVert_{2}} < \max_{\xi_s(t)}\frac{\left\lVert \tau_s^{{DFM}}(t)\right\rVert_{2}}{\left\lVert \xi_s(t)\right\rVert_{2}}
\end{equation}
This induced norm relationship satisfies the optimization goal in~\eqref{eq:optim}, thus, the problem in \eqref{eq:optim} can be reconsidered as a finding of Q filter bandwidth $\omega_Q$ which satisfies the condition in~\eqref{eq:opt}. {\jrev For brevity, the condition \eqref{eq:opt} can be re-arranged as}
\begin{equation}
\left\lVert (1-Q)K_s\right\rVert_{\infty} + \left\lVert sQHP_m^{-1}\right\rVert_{\infty} 
< \left\lVert K_s \right\rVert_{\infty}.
\label{eq:condition_simple}
\end{equation}
\begin{figure}
	\centering
	{\includegraphics[width=1\columnwidth]{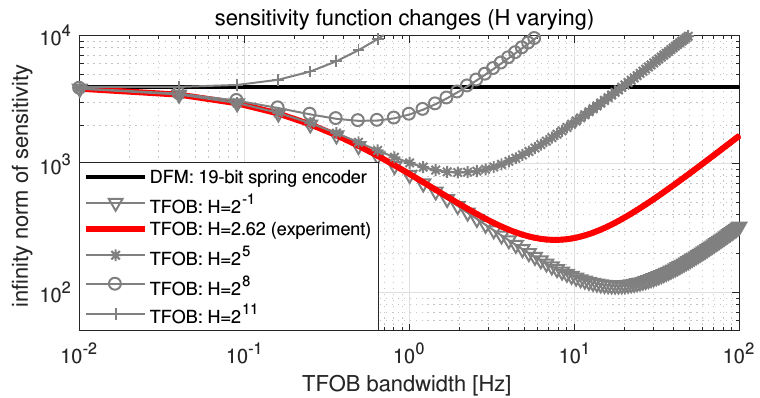}}
	\caption{Comparison of sensitivity against measurement error with {\jrev TFOB.}}
	\label{fig:H_inf}
	\vspace{-1em}
\end{figure}
{\crev Fig.}~\ref{fig:H_inf} shows the comparison of two infinity norms  $\left\lVert (1-Q)K_s\right\rVert_{\infty} + \left\lVert sQHP_m^{-1}\right\rVert_{\infty}$  and $\left\lVert K_s \right\rVert_{\infty}$  in~\eqref{eq:condition_simple} with regard to the bandwidth of Q filter in the $x$-axis {\orev to examine how the measurement error condition changes the magnitude. The} parameters, required for the comparison, are from Table~\ref{tab:parameters}, which is same as the experimental setup in Fig.~\ref{fig:hardware}. The black thick solid line represents $\left\lVert K_s \right\rVert_{\infty} $ which corresponds to the infinity norm of the DFM-based control sensitivity function in~\eqref{eq:error_DFM}, and other {\orev marked} lines {\jrev represent} 
$\left\lVert (1-Q)K_s\right\rVert_{\infty} + \left\lVert sQHP_m^{-1}\right\rVert_{\infty}$  which corresponds to the infinity norm of the TFOB-based control sensitivity function in~\eqref{eq:error_TFOB}. {\orev Various $H$ levels are also considered in Fig.~\ref{fig:H_inf}, from $2^{-1}$ to $2^{11}$.}

{\jrev Interestingly, the comparative result indicates that} 
that there is an optimal frequency bandwidth which minimizes $\left\lVert (1-Q)K_s\right\rVert_{\infty} + \left\lVert sQHP_m^{-1}\right\rVert_{\infty}$ such that TFOB-based control achieves the best performance. The effect of $H$ on the performance also can be analyzed using Fig.~\ref{fig:H_inf}; the smaller $H$ is, the better performance TFOB-based control can achieve; the bandwidth of TFOB value can be set higher when the motor-side encoder exhibits better measurement characteristic so that the performance of TFOB-based control can be improved.

As explained above, one simple interpretation of $H$ is the ratio of resolutions of two encoders taking the gear ratio also into consideration. With this interpretation, $H$ value of the experimental set up in Fig.~\ref{fig:hardware} is calculated as {\jrev follows:
	\[
	H = \frac{\mbox{(resolution of the motor encoder)}\times\mbox{(gear ratio)}} { \mbox{(resolution of the spring encoder)}} = 2.62.
	\]
	
	\noindent The case with $H= 2.62$} is depicted in Fig.~\ref{fig:hardware} (red thick line), where the optimal bandwidth is around {\jrev from 5-10~Hz}, which is to be verified in the following experiment.
\section{Experimental Verification}\label{sec:experiment}
The performance of the proposed TFOB-based force controller is verified through experiments in this section. At first, the experimental set up equipped with {\jrev the} SEA is introduced, then the following points are {\jrev experimentally investigated: 
	\begin{enumerate}
		\item [1)] tracking performance of the TFOB-based force control,
		\item [2)] force estimation accuracy during the control,
		\item [3)] robustness against noise coming from the sensor, and
		\item [4)] performance change with bandwidths of the Q filter.
	\end{enumerate}
	These performance and robustness of the proposed TFOB-based force control are compared with the conventional DFM-based control both in the time domain and the frequency domain.}
\begin{table}[t!]
	\centering
	\caption{Physical Parameters of the SEA in the Experiment Setup}
	\begin{tabular}{l l l l}
		\hline \hline
		notation &parameter& value&unit\\
		\hline
		$J_m$&motor inertia & 0.0000625&kgm$^2$\\
		$B_m$&motor damping & 0.0001023&Nms/rad\\
		$J_l$&load inertia & 0.216&kgm$^2$\\
		$J_l$&load damping & 0.0005&Nms/rad\\
		$K_s$&spring stiffness & 4950&Nm/rad\\
		$N$ & gear ratio& 100&-\\
		$K_p$& proportional gain & 1&-\\
		$K_d$& derivative gain & 0.014&-\\
		$\omega_Q$& bandwidth of Q(s)& from {\jrev 0.1-10}&Hz\\
		$\xi_m$ & motor encoder resolution&2000 {\jrev(x4)}&CPT\\
		$\xi_s$ & spring encoder resolution &19&bit\\
		$H$ & error ratio &2.621&-\\		
		\hline \hline				
	\end{tabular}
	\label{tab:parameters}
	\vspace{-1em}
\end{table}
\subsection{Experimental Setup}\label{sec:setup}
\subsubsection{Hardware Description}
{\crev Fig.}~\ref{fig:hardware} illustrates the experimental setup consisting of the rotary type SEA (Fig.~\ref{fig:hardware}(a)), the load and the torque sensor. The torque sensor (TNT-200 from Transducer Techniques with 22.6 Nm capacity) can directly measure the SEA torque output transmitted to the load.

Compact Planetary-geared Elastic Actuator (cPEA) is utilized in this experiment, the detailed structure and operating principle of which are presented in~\cite{7793816}. The physical parameters of the SEA are shown in Table~\ref{tab:parameters}.
\subsubsection{Experimental Protocol}
A proportional-derivative (PD) controller is adopted as the common feedback controller $C_f(s)$ for both DFB- and TFOB-based force control. The performances of two approaches are compared in three ways; 1) reference tracking error comparison with a step-wise reference, 2) deformation measurement error (noise) rejection performance comparison and 3) reference tracking performance comparison with various sinusoidal reference signals. Moreover, various Q filter bandwidths are tested for the TFOB-based force control experiments.
\subsection{Time Domain Experimental Results} \label{sec:time_results}
\subsubsection{Reference Tracking Performance}
A step signal is applied as the reference for both DFM- and TFOB-based force control. The tracking performance is evaluated by the difference between this reference and the actual SEA output torque, measured by a torque sensor.

\begin{figure}[t!]
	\centering
	\subfloat[]{\includegraphics[width=0.499\columnwidth]{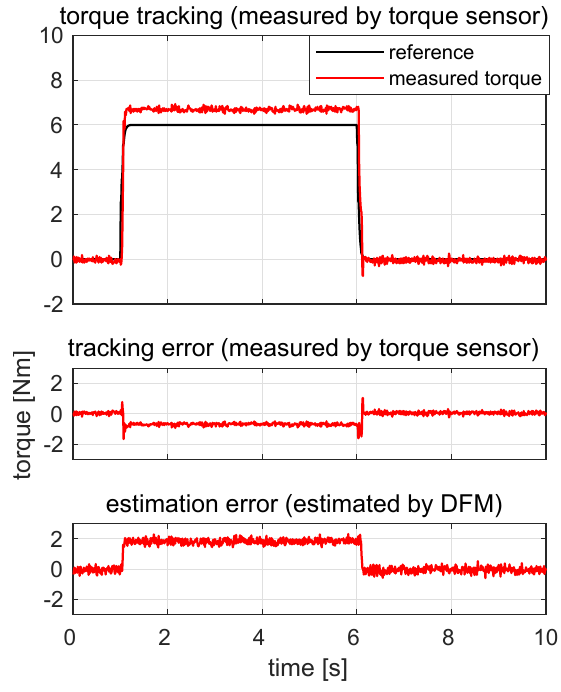}}~
	\subfloat[]{\includegraphics[width=0.499\columnwidth]{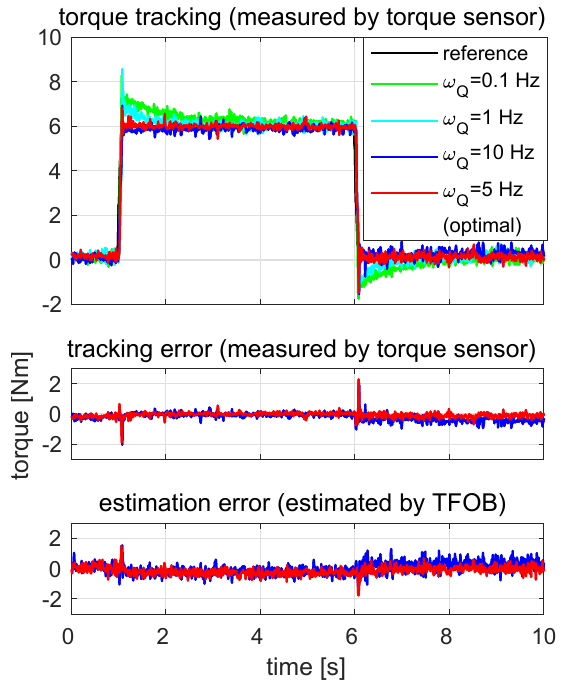}}
	\caption{Time domain reference tracking results of (a) DFM- and (b) TFOB-based force {\jrev control.}  }
	\label{fig:time}
	\vspace{-1em}
\end{figure}

{\crev Fig.}~\ref{fig:time} shows the results of reference tracking, where 6~Nm step-wise reference is given from 1~s to 6~s; {\crev Fig.}~\ref{fig:time}(a) is the result of DFM-based control, and Fig.~\ref{fig:time}(b) is the results of TFOB-based controls.
\begin{figure}[t!]
	\centering
	\subfloat[]{\includegraphics[width=0.499\columnwidth]{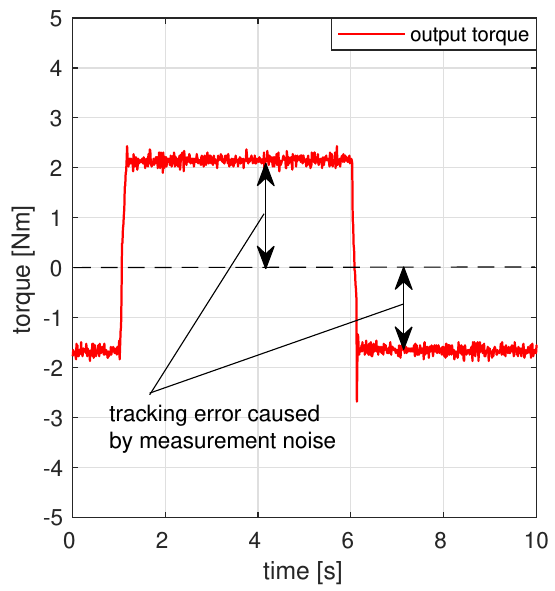}}~
	\subfloat[]{\includegraphics[width=0.499\columnwidth]{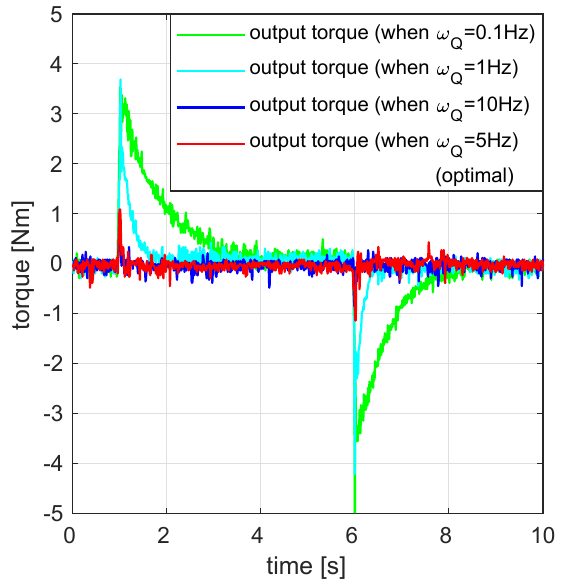}}
	\caption{Spring deflection measurement error rejection results of (a) DFM- and (b) TFOB-based {\jrev control.} }
	\label{fig:noise_s}
	\vspace{-1em}
\end{figure}
The output of the DFM-based control case shown in Fig.~\ref{fig:time}(a) exhibits steady state error which is more than 0.6~Nm (10\% of the reference).

On the contrary, TFOB-based force control case in Fig.~\ref{fig:time}(b) shows no steady state error regardless of how the reference signal changes. Further investigation with different bandwidth of Q filter verifies that higher $\omega_Q$ (5~Hz in this case) improves the tracking performance reducing overshoots.


\subsubsection{Noise Rejection Performance}
To verify the robustness against measurement errors, an additional step-wise measurement noise {\orev ($\xi_s^{\mbox{\footnotesize step}}$, $\xi_m^{\mbox{\footnotesize step}}$)} is added to the actual measurements $\theta_m$ of motor angle and spring deformation $\theta_s$, and the control performance is examined under the noise to investigate the robustness of each controller against this measurement error. 
\begin{figure}[t!]
	\centering
	\subfloat[]{\includegraphics[width=0.69\columnwidth]{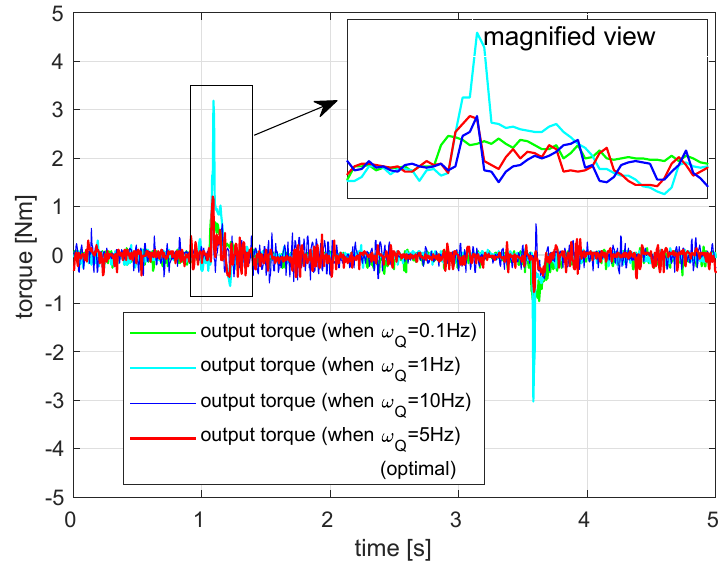}}
	\subfloat[]{\includegraphics[width=0.31\columnwidth]{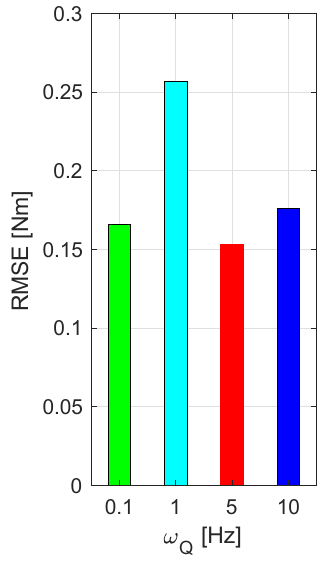}}
	\caption{Motor position measurement error rejection performance TFOB-based control: (a) time domain results and (b) RMSE {\jrev values.}}
	\vspace{-1em}
	\label{fig:noise_m}
\end{figure}


{\crev Fig.}~\ref{fig:noise_s} shows the actual output torque, which is supposed to be kept 0 by DFM-based control (Fig.~\ref{fig:noise_s}(a)) or TFOB-based control (Fig.~\ref{fig:noise_s}(b)) with 0 reference. A step-wise measurement error {\orev $\xi_s^{\mbox{\footnotesize step}}$} of 0.0005~rad is added to the spring measurement $\theta_s$ from 1~s to 5~s. The DFM-based control result in Fig.~\ref{fig:noise_s}(a) shows that the torque output is significantly affected by the noise, while the TFOB-based control in Fig.~\ref{fig:noise_s}(b) shows little error against the noise. The attenuation of the torque error against the measurement error depends on the $\omega_Q$, and Fig.~\ref{fig:noise_s}(b) verifies that the higher $\omega_Q$ rejects the effect of the measurement noise better.

{\crev Fig.}~\ref{fig:noise_m}(a) illustrates regulation performance of TFOB-based control in time domain when a step-wise error {\orev $\xi_m^{\mbox{\footnotesize step}}$} is added to the motor angle measurement $\theta_m$. Even though the measured torque output is affected by the noise, it is shown that higher $\omega_Q$ can remove the effect of the noise effectively. However, too high $\omega_Q$ (10~Hz, in this experiments) induces chattering in the output, which deteriorates the control performance of SEA. RMSEs are calculated to compare the performances with different $\omega_Q$, and displayed in Fig~\ref{fig:noise_m}(b). The RMSE comparison in Fig~\ref{fig:noise_m}(b) shows that $\omega_Q$=5~Hz shows the best regulation performance.

\vspace{-0.8em}
\subsection{Tracking Performance in the Frequency Domain} \label{sec:freq_results}
In this experiment, sinusoidal signals with various frequencies are added to the SEA torque control as the reference, and the RMSE values are calculated for each frequency. The tracking performance in the frequency domain is calculated in this way, and the results with DFM- and TFOB-based control with 4~different $\omega_Q$ setting are compared. 

\begin{figure}[t]
	\centering
	\subfloat[]{\includegraphics[width=0.499\columnwidth]{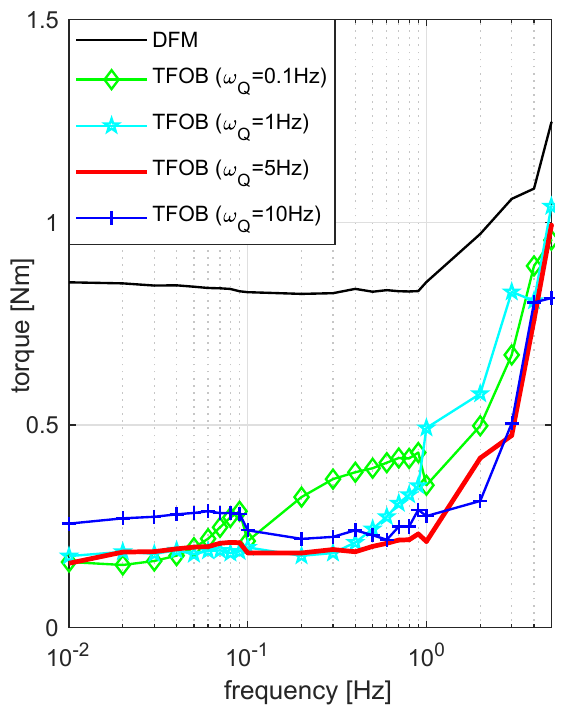}}~
	\subfloat[]{\includegraphics[width=0.499\columnwidth]{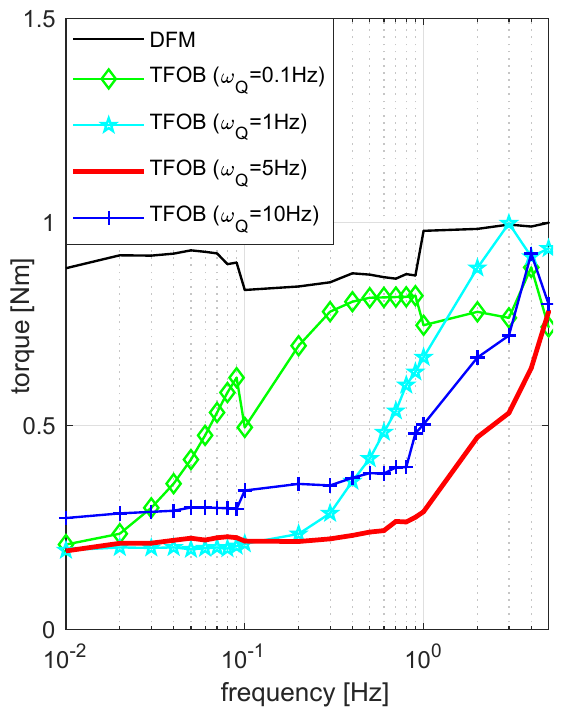}}
	\caption{Frequency domain results of (a) tracking errors and  (b) estimation errors under the DFM- and TFOB-based {\jrev control.}}
	\label{fig:freq_r}
	\vspace{-1em}
\end{figure}

{\crev Fig.}~\ref{fig:freq_r} shows the result where each dot represents the RMSE at each frequency (the magnitude of the sinusoidal reference is set to 3~Nm). The tracking errors are compared in Fig.~\ref{fig:freq_r}(a), and the estimation errors are compared in Fig.~\ref{fig:freq_r}(b), and the results validate that the proposed TFOB-based control can improve the performance compared with the DFM-based control at all frequencies. 

%
\section{Conclusion}
This paper proposed the accurate force control algorithm for SEA systems based on the TFOB.
The contributions of this paper is concluded as {\jrev follows:
	first, the
	causes of SEA force measurement issues are modeled and analyzed in the viewpoint of deterministic and stochastic behavior. The results indicated that the errors characteristic of SEA force measurement is Gaussian; second, 
	TFOB was developed to achieve precise force measurement with consideration of errors for motor-side position and spring deformation measurement. The verifications of TFOB-based force estimation performance showed that the optimal performance can be found at the specific frequency bandwidth of TFOB;    
	third, TFOB-based force controller was designed to accurately control the output force of SEA. In addition, the tuning method for the controller was {\orev proposed} by using the dynamics of the closed-loop SEA system and the investigated Gaussian behavior of the measurement~error;
	and last, experiments have been performed to explore the error behavior, and to verify the precision of proposed TFOB-based force controller and the observer.}

\bibliography{TFOB_based_control_final_draft}{}

\begin{thebibliography}{10}

\bibitem{lee2017generalization}
C.~Lee, S.~Kwak, J.~Kwak, and S.~Oh, ``Generalization of series elastic
  actuator configurations and dynamic behavior comparison,'' in {\em
  Actuators}, vol.~6, p.~26, MDPI, 2017.

\bibitem{7579567}
S.~Oh and K.~Kong, ``High-precision robust force control of a series elastic
  actuator,'' {\em {IEEE/ASME} Trans. Mechatronics}, vol.~22, pp.~71--80, Feb
  2017.

\bibitem{tsagarakis2017walk}
N.~G. Tsagarakis, D.~G. Caldwell, F.~Negrello, W.~Choi, L.~Baccelliere, V.~Loc,
  J.~Noorden, L.~Muratore, A.~Margan, A.~Cardellino, {\em et~al.}, ``Walk-man:
  A high-performance humanoid platform for realistic environments,'' {\em
  Journal of Field Robotics}, vol.~34, no.~7, pp.~1225--1259, 2017.

\bibitem{491410}
K.~Ohnishi, M.~Shibata, and T.~Murakami, ``Motion control for advanced
  mechatronics,'' {\em {IEEE/ASME} Trans. Mechatronics}, vol.~1, pp.~56--67,
  March 1996.

\bibitem{negrello2017design}
F.~Negrello, M.~Catalano, M.~Garabini, M.~Poggiani, D.~Caldwell, N.~Tsagarakis,
  and A.~Bicchi, ``Design and characterization of a novel high-compliance
  spring for robots with soft joints,'' in {\em 2017 IEEE International
  Conference on Advanced Intelligent Mechatronics (AIM)}, pp.~271--278, IEEE,
  2017.

\bibitem{wang2015novel}
M.~Wang, L.~Sun, W.~Yin, S.~Dong, and J.~Liu, ``A novel sliding mode control
  for series elastic actuator torque tracking with an extended disturbance
  observer,'' in {\em 2015 IEEE International Conference on Robotics and
  Biomimetics (ROBIO)}, pp.~2407--2412, IEEE, 2015.

\bibitem{wang2014output}
W.~Wang and S.~Sugano, ``Output torque regulation through series elastic
  actuation with torsion spring hysteresis,'' in {\em 2014 IEEE International
  Conference on Robotics and Biomimetics (ROBIO)}, pp.~701--706, IEEE, 2014.

\bibitem{choi2017low}
W.~Choi, J.~Won, J.~Lee, and J.~Park, ``Low stiffness design and hysteresis
  compensation torque control of sea for active exercise rehabilitation
  robots,'' {\em Autonomous Robots}, vol.~41, no.~5, pp.~1221--1242, 2017.

\bibitem{makarov2016modeling}
M.~Makarov, M.~Grossard, P.~Rodr{\'\i}guez-Ayerbe, and D.~Dumur, ``Modeling and
  preview $\text{H}_\infty$ control design for motion control of elastic-joint
  robots with uncertainties,'' {\em Trans. Ind. Electron.}, vol.~63, no.~10,
  pp.~6429--6438, 2016.

\bibitem{austin2015control}
J.~Austin, A.~Schepelmann, and H.~Geyer, ``Control and evaluation of series
  elastic actuators with nonlinear rubber springs,'' in {\em 2015 IEEE/RSJ
  International Conference on Intelligent Robots and Systems (IROS)},
  pp.~6563--6568, IEEE, 2015.

\bibitem{ruderman2016compensation}
M.~Ruderman, ``Compensation of nonlinear torsion in flexible joint robots:
  Comparison of two approaches,'' {\em {IEEE} Trans. Ind. Electron.}, vol.~63,
  no.~9, pp.~5744--5751, 2016.

\bibitem{lee2017two}
C.~Lee, J.~Lee, J.~Malzahn, N.~Tsagarakis, and S.~Oh, ``A two-staged residual
  for resilient external torque estimation with series elastic actuators,'' in
  {\em Humanoid Robotics (Humanoids), 2017 IEEE-RAS 17th International
  Conference on}, pp.~817--823, IEEE, 2017.

\bibitem{mitsantisuk2013design}
C.~Mitsantisuk, M.~Nandayapa, K.~Ohishi, and S.~Katsura, ``Design for
  sensorless force control of flexible robot by using resonance ratio control
  based on coefficient diagram method,'' {\em automatika}, vol.~54, no.~1,
  pp.~62--73, 2013.

\bibitem{lee2018residual}
J.~Lee, C.~Lee, N.~Tsagarakis, and S.~Oh, ``Residual-based external torque
  estimation in series elastic actuators over a wide stiffness range: Frequency
  domain approach,'' {\em IEEE Robotics and Automation Letters}, vol.~3, no.~3,
  pp.~1442--1449, 2018.

\bibitem{yamada2018proposal}
S.~Yamada and H.~Fujimoto, ``Proposal of state-dependent minimum variance
  estimation of load-side external torque considering modeling and measurement
  errors,'' in {\em 2018 IEEE 27th International Symposium on Industrial
  Electronics (ISIE)}, pp.~1--6, IEEE, 2018.

\bibitem{ruderman2016sensorless}
M.~Ruderman and M.~Iwasaki, ``Sensorless torsion control of elastic-joint
  robots with hysteresis and friction,'' {\em {IEEE} Trans. Ind. Electron.},
  vol.~63, no.~3, pp.~1889--1899, 2016.

\bibitem{haddadin2017robot}
S.~Haddadin, A.~De~Luca, and A.~Albu-Sch{\"a}ffer, ``Robot collisions: A survey
  on detection, isolation, and identification,'' {\em {IEEE} Trans. Robot.},
  vol.~33, no.~6, pp.~1292--1312, 2017.

\bibitem{murakami1993torque}
T.~Murakami, F.~Yu, and K.~Ohnishi, ``Torque sensorless control in
  multidegree-of-freedom manipulator,'' {\em {IEEE} Trans. Ind. Electron.},
  vol.~40, no.~2, pp.~259--265, 1993.

\bibitem{oh2014design}
S.~Oh, K.~Kong, and Y.~Hori, ``Design and analysis of force-sensor-less
  power-assist control,'' {\em {IEEE} Trans. Ind. Electron.}, vol.~61, no.~2,
  pp.~985--993, 2014.

\bibitem{8371176}
C.~Lee and S.~Oh, ``Integrated transmission force estimation method for series
  elastic actuators,'' in {\em 2018 IEEE 15th International Workshop on
  Advanced Motion Control (AMC)}, pp.~681--686, March 2018.

\bibitem{sivashankar1992induced}
N.~Sivashankar and P.~P. Khargonekar, ``Induced norms for sampled-data
  systems,'' {\em Automatica}, vol.~28, no.~6, pp.~1267--1272, 1992.

\bibitem{7793816}
C.~Lee and S.~Oh, ``Configuration and performance analysis of a compact
  planetary geared elastic actuator,'' in {\em IECON 2016 - 42nd Annual
  Conference of the IEEE Industrial Electronics Society}, pp.~6391--6396, Oct
  2016.

\end{thebibliography}
\bibliographystyle{ieeetr}
\vspace{-1px}

\end{document}